\documentclass[10pt, a4paper]{article}

\usepackage[]{lrec-coling2024} 

\newcommand{\R}{\mathbb{R}}
\newcommand{\blfootnote}[1]{%
    \begingroup
    \renewcommand\thefootnote{}\footnote{#1}%
    \addtocounter{footnote}{-1}%
    \endgroup
}

\usepackage{amsfonts}
\usepackage{amsmath}
\usepackage{svg}
\usepackage{tikz}
\usepackage{enumitem}

\title{\textbf{Detecting Hallucination and Coverage Errors in Retrieval Augmented Generation for Controversial Topics}}

\name{Tyler A. Chang*$^{1,2}$, Katrin Tomanek*$^1$, Jessica Hoffmann$^1$, Nithum Thain$^1$, \\
{\bf \large  Erin van Liemt$^1$, Kathleen Meier-Hellstern$^{1\diamondsuit}$, Lucas Dixon$^{1\diamondsuit}$}\vspace{0.2cm}}

\address{$^1$Google Research \\
$^2$UC San Diego \\
tachang@ucsd.edu \\
\{katrintomanek, jhoffmann, nthain, evanliemt, kathyhellstern, ldixon\}@google.com\\}

\abstract{
We explore a strategy to handle controversial topics in LLM-based chatbots based on Wikipedia’s Neutral Point of View (NPOV) principle: acknowledge the absence of a single true answer and surface multiple perspectives. We frame this as retrieval augmented generation, where perspectives are retrieved from a knowledge base and the LLM is tasked with generating a fluent and faithful response from the given perspectives. As a starting point, we use a deterministic retrieval system and then focus on common LLM failure modes that arise during this approach to text generation, namely hallucination and coverage errors. We propose and evaluate three methods to detect such errors based on (1) word-overlap, (2) salience, and (3) LLM-based classifiers. Our results demonstrate that LLM-based classifiers, even when trained only on synthetic errors, achieve high error detection performance, with ROC AUC scores of 95.3\% for hallucination and 90.5\% for coverage error detection on unambiguous error cases. We show that when no training data is available, our other methods still yield good results on hallucination (84.0\%) and coverage error (85.2\%) detection.
\\ \newline \Keywords{conversational systems, natural language generation, evaluation methodologies} }
\begin{document}

\maketitleabstract

\section{Introduction}
\renewcommand{\arraystretch}{1.1}  
Large\blfootnote{*Joint first authorship.}\blfootnote{$^{\diamondsuit}$Research group leadership.} Language Models (LLMs) have achieved state-of-the-art performance on a wide range of tasks, and a growing audience of users is engaging with LLM-driven chatbots.\footnote{Among others: \href{https://openai.com/blog/chatgpt}{https://openai.com/blog/chatgpt}; \href{https://bard.google.com}{https://bard.google.com}; \href{https://www.anthropic.com/index/introducing-claude}{https://www.anthropic.com/\\index/introducing-claude}.} While these chatbots are highly flexible and generalizable, they are known to struggle with factuality and bias \cite{sheng-etal-2019-woman,shuster-etal-2021-retrieval-augmentation,chang-bergen-2023-language}. In many real world scenarios, model developers require more precise control over LLM-based chatbot responses.

In this paper, we investigate how LLMs can be used with retrieval augmented generation for controversial topics, and we propose methods to detect errors in the tuned LLM responses.
In retrieval augmented generation, factual information is retrieved and provided as additional context to an LLM \citep{lewis-retrieval-2020,li2022survey,azure-2023-retrieval,iyer-thallam-2023-building}.
Through curated retrieval sources, retrieval augmented generation enables fine-grained control over LLM responses.
However, in the case of controversial topics, users often seek information for which there are not agreed-upon factual answers. These topics range from the inconsequential (e.g. ``the superiority of the Yankees vs. the Red Sox'') to the fundamental (``What religious faith should I adhere to?''). Building useful LLMs requires the ability to ensure that LLM responses adhere to desired levels of neutrality and nuance in such cases.

Thus, we introduce the \textbf{NPOV Response Task}: given a query about a controversial topic, the model retrieves arguments for multiple perspectives and is tasked to generate a multi-perspective response, inspired by Wikipedia's Neutral Point of View (NPOV) principle. We use a deterministic argument retrieval system, and we focus on the challenge of faithful response generation from provided arguments. We adapt a conversational LLM to this task and examine two common error types that violate faithfulness to inputs: (1) \textbf{hallucinations} (generating unprovided arguments), and (2) \textbf{coverage errors} (omitting provided arguments).

We build a dataset of model query-response pairs, conditioned on arguments from Britannica's ProCon \cite{procon}.
Using  expert annotators, we identify instances of hallucination and coverage errors. We then propose methods for detecting such hallucination and coverage errors, both with and without access to human-labeled data.

\setlength{\belowcaptionskip}{-0.25cm}
\begin{figure*}[t!]
    \centering
    \includegraphics[width=1.0\linewidth]{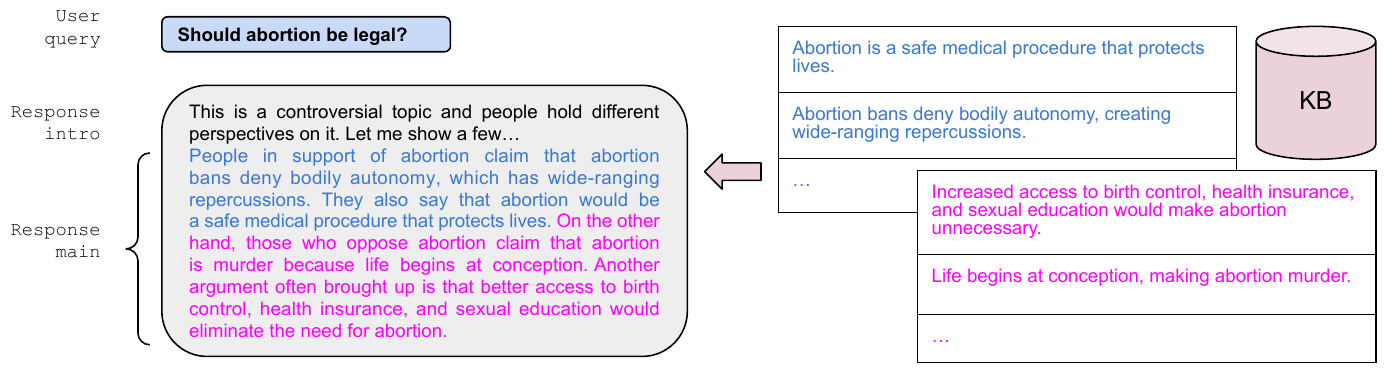}
    \caption{Example NPOV response to a user query on a senstive topic (left) based on pro and con perspectives with two arguments each, as obtained from a knowledge base (right). Arguments taken from \citet{procon}. Our error detection methods focus on the NPOV main response.}
    \label{fig:nuanced_response_generation_overview}
\end{figure*}
\setlength{\belowcaptionskip}{-0.0cm}

Our main results demonstrate that with access to error-free examples and examples containing only synthetic errors, LLM-based classifiers can achieve ROC AUCs of 95.3\% and 90.5\% in detecting organic hallucinations and coverage errors respectively on our task. Even without access to annotated data, we can leverage salience and word overlap techniques to achieve ROC AUCs of 84.0\% for hallucinations and 85.2\% for coverage errors.
While we focus on NPOV response generation, our approaches can be applied more generally to detect hallucination and coverage errors in retrieval augmented generation, facilitating finer-grained control over LLM responses.

\section{Handling Controversial Topics}
\label{sec:npov_response}

Our work is centered around how LLMs can be controlled to respond to queries about controversial topics for which there is no single correct answer. For example, in response to ``Should abortion be legal?'', an LLM without direction might produce a highly opinionated or offensive response.
To address such concerns, ``guardrails'' are often added to LLMs, either completely preventing the generation of responses to such topics or responding with canned answers (``I am just a language model and cannot answer this question...''). Such approaches can lead to erasure harm and reduce the usefulness of the system on potentially important topics. Another approach is to personalize responses to align with a user's position; however, this can reinforce harmful biases and popular misconceptions, and act as a chatbot echo chamber.

As an alternative strategy, we propose to acknowledge the lack of agreement and surface main viewpoints instead. This approach is inspired by Wikipedia's \textbf{Neutral Point of View (NPOV)} principle, which requires that content is written such that it represents ``fairly, proportionately, and, as far as possible, without editorial bias, all the significant views that have been published by reliable sources on a topic.''\footnote{From \href{https://en.wikipedia.org/wiki/Wikipedia:Neutral_point_of_view}{https://en.wikipedia.org/wiki/Wikipedia:Neutral\\\_point\_of\_view}, last accessed 2023/10/20.}
Figure~\ref{fig:nuanced_response_generation_overview} (left) gives an example of an NPOV response on a highly controversial topic. We explore whether such responses can be generated by an LLM using retrieval augmented generation, and we detect common failure modes such as hallucination and coverage errors.

\subsection{NPOV Response Generator}
\label{ss:npov_writer}

We separate \textit{response} generation from \textit{content} generation. For the scope of this paper, we assume that there is a content retrieval process and a knowledge base of curated arguments for different perspectives. The knowledge base we use in this paper consists of arguments from Britannica's ProCon website (\S\ref{sec:procon}). 

The NPOV Response Task is then: given the user query and retrieved perspectives (where perspectives consist of concatenated arguments), generate a response that consists of an introduction sentence, serving as a bridge from the user query, and a verbalization of the given perspectives.
When generating the response, relevant aspects of the given arguments must not be dropped (ensure full coverage) and no other arguments should be added (avoid hallucinations).
This task formulation gives model developers fine-grained control over LLM responses.
An example is shown in Figure~\ref{fig:nuanced_response_generation_overview}.

We use soft prompt-tuning~\cite{lester2021} to adapt an LLM to generate NPOV responses given pro and con arguments. Our base LLM is a 64B decoder-only LaMDA model pre-trained on public dialog data and web text \citep{lamda-paper}.
We use a soft prompt length of 5 tokens, and we train for 20K steps with batch size 16 and learning rate 0.1.
We typically reach maximum dev set performance after 2-5K steps.
Specific prompt format and detailed hyperparameters are in Appendix \ref{app:prompt-tuning}.

Our training set consists of 80 query-response pairs covering 9 controversial topics from ProCon (\S\ref{sec:procon}). ProCon question headers (e.g. ``Should abortion be legal?'') are used as user queries. For each topic, we randomly sample one, two, or three arguments from both the pro and con side in ProCon\footnote{We always ensure the same number of pro and con arguments.} and then manually write several paraphrased responses capturing these arguments. 
We observe that after prompt-tuning, the NPOV Response Generator generalizes well beyond the topics and arguments seen during training.

\subsection{ProCon as a Knowledge Base}
\label{sec:procon}

Britannica's ProCon \cite{procon} is a website presenting pros and cons for commonly debated topics. Pros and cons are researched and compiled by ProCon research staff and editors, and they aim to be nonpartisan.\footnote{Of course, not all controversial topics can be framed as pro versus con debates, and such a binary framing of highly complex topics can omit important nuance (see Ethical Considerations).}
As of October 2022, ProCon contains 72 active (i.e. ``non-archived'') topics.
For both the pro and con perspective for each topic, several arguments are given, each consisting of a short argument phrase accompanied by a longer explanation.
The median number of arguments per perspective per topic is 4, but some topics contain many more arguments (e.g. \textit{Social Media} has 23 arguments per perspective).
We randomly sample ProCon arguments as inputs to the NPOV Response Generator for each topic (\S\ref{sec:annotation-procedure}).
Each topic is associated with a leading question in ProCon (e.g. ``Should abortion be legal?''), which we treat as the user query asked to the LLM.

\section{Methods to Detect Hallucinations and Coverage Errors}
\label{sec:methods}
We focus on hallucination and coverage error detection, adopting the following definitions:
\begin{itemize}[leftmargin=0.5cm]
    \setlength\itemsep{0.0em}
    \item If the generated response contains at least one argument which was not provided, we call this a \textbf{hallucination}.
    \item If one or more of the given arguments is completely dropped from the response, we call this a \textbf{coverage error}.
\end{itemize}
We call these \textbf{full errors}, as they address the hallucination or coverage of a full argument.
On top of these well-defined errors, we notice that the NPOV Response Generator sometimes produces other unfaithful changes to arguments, including: (1) partial hallucinations (slight meaning change, e.g.``consensus'' becomes ``unanimity''), (2) partial coverage errors (only a part of the argument is dropped), (3) repetitions (response contains the same given argument multiple times), and (4) perspective confusions (response inverts the perspectives, e.g. pro arguments are presented as cons). We call all of these \textbf{ambiguous errors}.

We propose three methods for detecting hallucination and coverage errors in generated responses: ROUGE, salience, and LLM-based classifiers.

\subsection{ROUGE}
As a baseline, we use ROUGE-1 (word-matching) to compute hallucination and coverage error scores \cite{lin-2004-rouge}. For a given response from the NPOV Response Generator, ROUGE calculates the proportion of \textit{response words} that are matched in the \textit{input arguments} (ROUGE-1 precision) and the proportion of \textit{input argument words} that are matched in the \textit{response} (ROUGE-1 recall).\footnote{We also implemented a hallucination and coverage error detection method that matched input and response arguments with BERTScore \citep{zhang2020bertscore}, but we obtained similar results to ROUGE. We omit results due to space limitations.}
Low precision is indicative of hallucination, and low recall is indicative of a coverage error.
Because the NPOV Response Task requires that both input perspectives be covered, we compute ROUGE-1 recall separately for each input perspective and then compute the minimum as our overall recall score.
For ROUGE, words are defined using whitespace and punctuation separation, dropping stop words and using word stemming from NLTK \cite{nltk}.

\subsection{Salience}
\label{sec:salience}
Aside from word matching, previous work has proposed methods to attribute output subword tokens to input tokens in LLMs using model gradients \citep{denil-2014-extraction,li-etal-2016-visualizing,bastings-filippova-2020-elephant}.
These methods are computationally costly, but they can often capture more nuance (e.g. word synonyms and token interactions) than simple word-matching.
One popular approach is to compute the logit (pre-softmax probability) gradient for each output token with respect to each input token embedding, producing a gradient vector for each input-output token pair.
The attribution from each input to the output token is defined as the dot product between the corresponding gradient vector and the input token embedding \citep{denil-2014-extraction}.\footnote{We obtain similar results using gradient L2 norms.}

\setlength{\belowcaptionskip}{-0.25cm}
\setlength{\abovecaptionskip}{0.05cm}
\begin{figure*}[t]
    \begin{center}
    \includegraphics[width=0.65\linewidth]{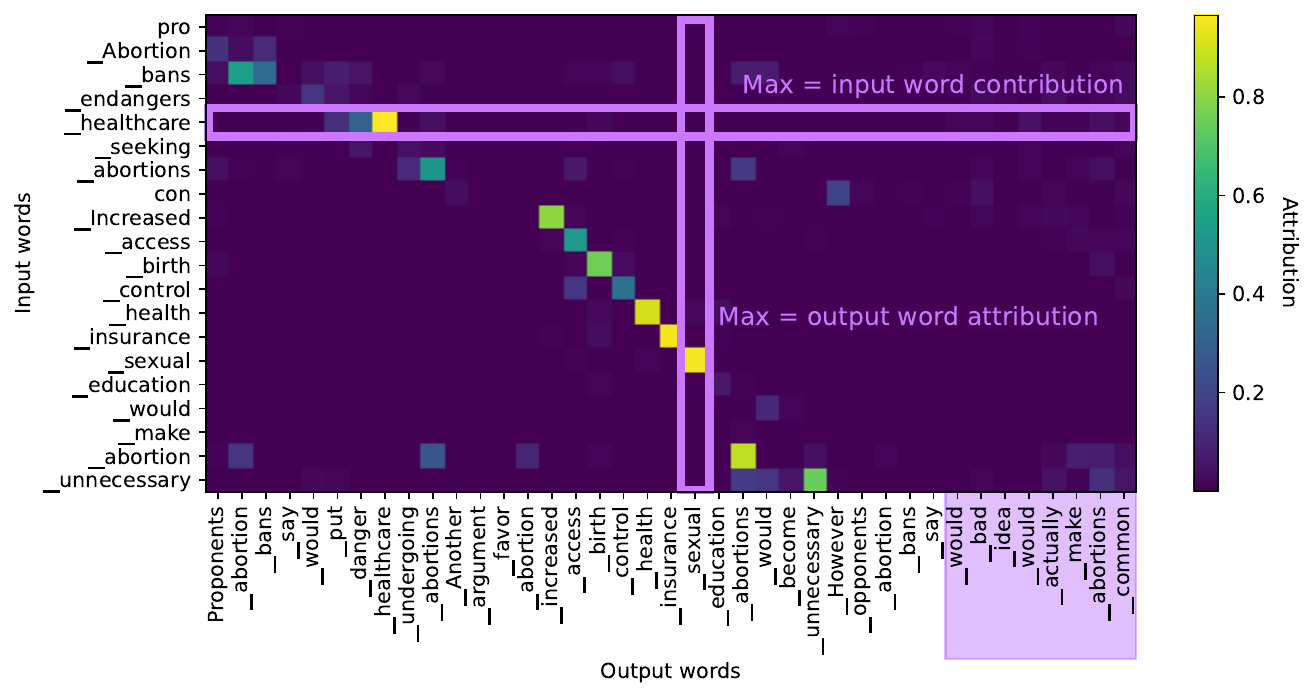} \newline
    \includegraphics[width=0.35\linewidth]{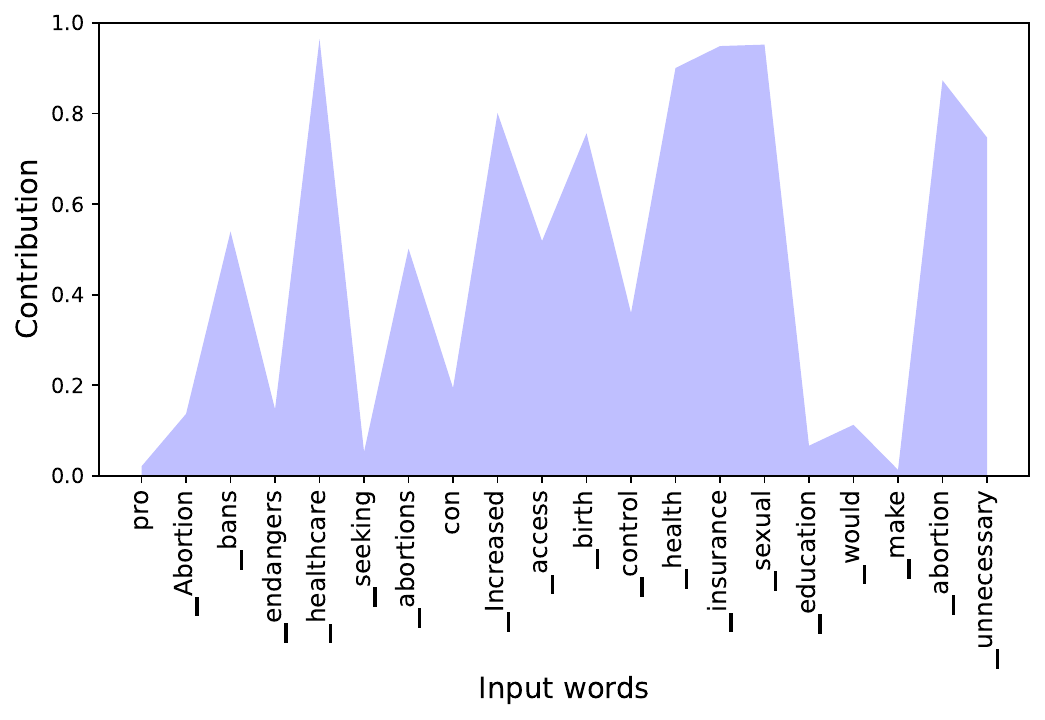} \hspace{0.5cm}
    \includegraphics[width=0.35\linewidth]{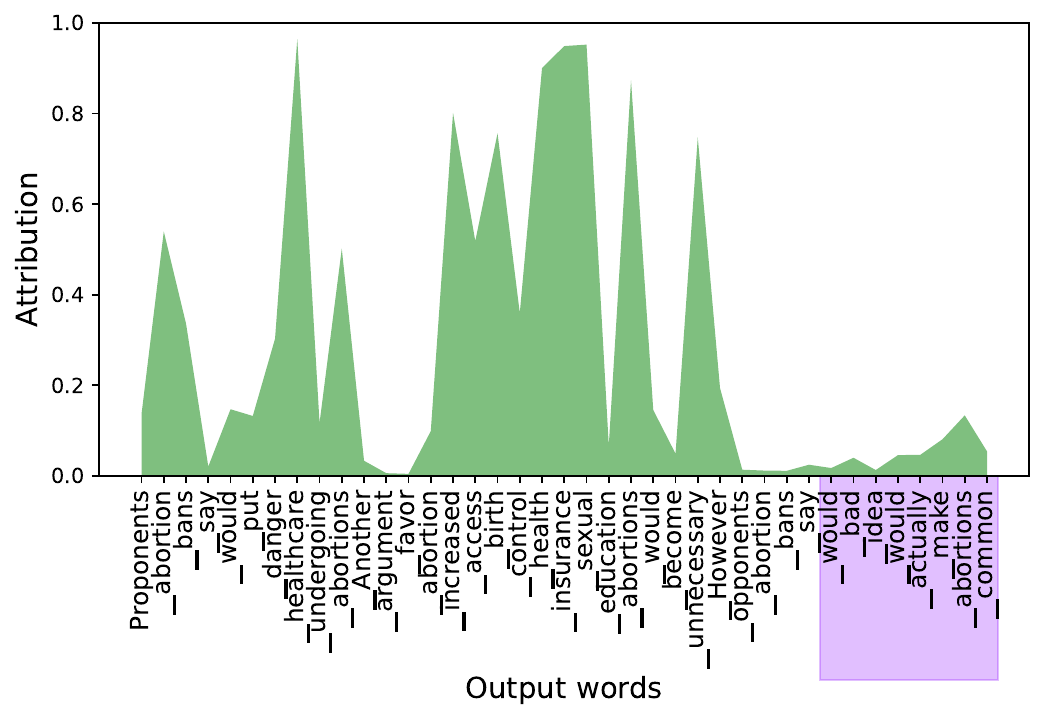}
    \end{center}
    \caption{Top: salience map from input argument content words (rows) to model response content words (columns). Bottom: individual word scores for contribution (input words; left) and attribution (response words; right). The purple highlighted words are hallucinated in the model response.}
    \label{fig:salience-map}
\end{figure*}
\setlength{\abovecaptionskip}{0.25cm}
\setlength{\belowcaptionskip}{0.0cm}

In the NPOV Response Generation scenario, there are attribution values from each input token (e.g. the given arguments per perspective and the user query) and each previously generated token to each output token.
This produces a token-to-token salience map $M_{\textrm{tokens}} \in \R^{(m+\ell) \times \ell}$, where $m$ is the number of input tokens and $\ell$ is the number of model response tokens.
Before any further processing, we square the salience map and normalize columns to sum to one (i.e. the attributions to each output token sum to one).

Because we are primarily concerned with hallucination and coverage errors for content words, we convert the subword token-to-token salience map to a word-to-word salience map $M_{\textrm{words}}$.
We define words by concatenating consecutive LLM tokens that are not separated by punctuation or whitespace; we then drop stop words, as defined in NLTK \cite{nltk}.
We define the attribution from an input word $w_0$ to an output word $w_1$ as the maximum attribution from any subword token in $w_0$ to any subword token in $w_1$.
We restrict our salience maps to the input argument words (rows) and the output NPOV response words (columns).
A sample word-to-word salience map for a query-response pair is shown in Figure~\ref{fig:salience-map}.

Qualitatively, we observe that covered words tend to have a high contribution to a single corresponding word in the response.
Thus, we define the \textit{contribution} score of an input argument word as its maximum contribution to any response word (i.e. maximum for each row of $M_{\textrm{words}}$).
We define the \textit{attribution} score of a response word as its maximum attribution from any input argument word (i.e. maximum for each column of $M_{\textrm{words}}$).
Contribution and attribution scores for input words and response words respectively are shown in Figure~\ref{fig:salience-map}.

To compute an example-level \textit{contribution} score for a query-response pair, we compute the mean contribution score over words in each of the two input perspectives.
As with ROUGE, we take the minimum of the two perspective contributions as a final contribution score.
To compute an example-level \textit{attribution} score, we compute the mean attribution score over all response words.
Finally, hallucination and coverage error scores in $[0, 1]$ are computed by subtracting the attribution and contribution scores respectively from 1.0.
Formal equations are in Appendix \ref{app:salience-formulas}.

\subsection{LLM-Based Classifiers}
\label{ss:llm_classifiers}

The two previous methods for detecting hallucination and coverage errors are data-free, not requiring labeled model responses for training. For the non-data-free scenario, we explore how well LLM-based classifiers perform on these tasks, relying on a small set of human annotations of model responses ($\sim$500 examples; \S\ref{ss:classifier_ablation}).
Our classifiers are built on FLAN-PaLMChilla, a 62B decoder-only LLM \citep{palm-paper} which has been instruction-tuned on a large number of tasks \citep{chung2022scaling}.
We use soft prompt-tuning to adapt this LLM into classifiers for hallucination and coverage error detection.
The classifiers have as input: (1) the user query, (2) the generated NPOV response, and (3) the given arguments per perspective.
We train the LLM to predict the label ``NO'' if there is a full error and ``YES'' otherwise.
Prompt-tuning hyperparameters are the same as \S\ref{ss:npov_writer}; specific prompt formats are in Appendix \ref{app:prompt-tuning}.
We tune the classifiers separately for the two error types.

For inference, we generate error classification scores in $[0,1]$ by obtaining the LLM's log perplexity scores for the tokens corresponding to the two output class labels (``YES'' and ``NO''), apply softmax, and take the score of the negative class (``NO'').\footnote{For single-token labels, this score equals the probability of ``NO'' conditioned on either  ``YES'' or ``NO'' output. We obtain similar results training the models with flipped labels, i.e. ``YES'' for errors and ``NO'' otherwise.}

\section{Dataset}
\label{sec:dataset}

To train and evaluate the hallucination and coverage error detection methods above on the NPOV Response Task, we construct datasets of organic (i.e. naturally occurring) and synthetic errors, with and without paraphrasing.

\subsection{Annotation Procedure}
\label{sec:annotation-procedure}

For each of the 72 controversial topics from ProCon, we generate a unique query and up to 18 query-response pairs
by first randomly sampling combinations of pro and con arguments, with either 1, 2, or 3 arguments per side, and then using the NPOV Response Generator to generate a response.
We annotate these query-response pairs (also called \textit{examples}) in three stages to (1) identify error-free examples, (2) identify examples with errors, and (3) generate paraphrased examples:
\begin{enumerate}[leftmargin=0.5cm]
    \setlength\itemsep{0.0em}
    \item For the first three examples per topic, we sample two generator responses, with sampling temperatures 0.0 and 0.7.
    We annotate whether responses contain hallucinations or coverage errors, annotating examples with a mix of the two temperatures.
    We annotate the token spans in the response that cover each input argument, along with any hallucinated response spans and uncovered input argument spans.
    \item Because examples with hallucination and coverage errors are less frequent than error-free examples even for high temperatures (20.0\% errors in 0.7 temperature responses), we sample a single 0.7 temperature response for each of the remaining (up to) 15 examples per topic.\footnote{Preliminary experiments with the NPOV Response Generator suggest that temperatures above 1.0 tend to produce overly long and irrelevant responses.}
    We annotate for hallucination and coverage errors, including full and ambiguous errors (\S\ref{sec:methods}).
    \item Hallucination and coverage error detection methods should capture whether meaning is retained between input arguments and generated responses, even if the arguments are not copied verbatim. We therefore generate examples with enforced paraphrasing between the input arguments and the response. To do so, we paraphrase the input arguments for all error-free examples generated in Step 1. For each argument, we use an off-the-shelf paraphrasing tool and manually verify that the paraphrasing does not induce substantial meaning change.\footnote{We use \href{https://quillbot.com/}{https://quillbot.com/} for paraphrasing. We find it more efficient to paraphrase the input arguments than to paraphrase the whole response.}
\end{enumerate}
In total, we identify 160 examples with no errors and 326 examples with at least one error, and we generate 152 paraphrased examples with no errors.

\subsubsection{Inter-Annotator Agreement}
\label{ss:interannotator-agreement}

To validate the viability and coherence of our annotation task, we hired a team of 10 external annotators to re-identify both hallucination and coverage errors in our dataset.
Our annotation provider was paid 49 USD per hour for a total of 25 hours of work (Appendix \ref{app:annotations}). Annotators were presented with 188 of the query-response pairs annotated in annotation Step 1 (\S\ref{sec:annotation-procedure}) and 86 pairs from Step 2.
Given the user query, the provided arguments, and the response from the NPOV Response Generator, annotators were asked to mark whether each response had a hallucination or coverage error. Each query-response pair was annotated by 5 annotators.
We compare the annotator majority vote to our annotated labels, finding 90\% agreement for hallucinations and 94\% for coverage errors.
To measure inter-annotator agreement, we compute Krippendorff's alpha for hallucinations ($\alpha=0.60$) and coverage errors ($\alpha=0.73$) across the 10 annotators.
These values are in line with or above similar text classification tasks \cite{wulczyn2017exmachina}.

\subsection{Synthetic Errors Dataset}
\label{sec:synthetic-generation}

Due to the relative rarity of \textit{organic} errors produced by the NPOV Response Generator, we synthetically generate examples with errors by modifying error-free query-response pairs.
Specifically, we modify the list of given arguments while keeping the original response unchanged. For coverage errors, we add one randomly sampled unused argument for the given topic from ProCon and add it to the list of given arguments. This creates a full coverage error because the original response does not cover this argument. For hallucinations, we randomly remove one of the given arguments. This creates a hallucination because the original response still addresses the removed argument.
We apply synthetic error generation to both paraphrased and unparaphrased examples that were annotated as error-free in \S\ref{sec:annotation-procedure} (312 examples), generating 667 new examples with synthetic hallucinations, synthetic coverage errors, or both.

\setlength{\belowcaptionskip}{-0.2cm}
\begin{table*}[t]
    \centering
    \begin{tabular}{|l|c|c|c||c|c|c|c|c|c|}
        \cline{2-7}
        \multicolumn{1}{c|}{} & \multicolumn{3}{c||}{Hallucinations} & \multicolumn{3}{c|}{Coverage Errors} \\
        \hline
        Test set error type & ROUGE & Salience & Classifier & ROUGE & Salience & Classifier \\
        \hline
        \hline
        Full organic & 0.840 & 0.808 & \textbf{0.953} & 0.795 & 0.852 & \textbf{0.905} \\
        Unparaphrased synthetic & 0.772 & 0.736 & \textbf{0.998} & 0.890 & 0.875 & \textbf{0.986} \\        
        Paraphrased synthetic & 0.680 & 0.708 & \textbf{0.977} & 0.746 & 0.831 & \textbf{0.993} \\
        Ambiguous organic & 0.814 & 0.772 & \textbf{0.851} & \textbf{0.834} & 0.755 & 0.756 \\
        \hline
    \end{tabular}
    \caption{ROC AUCs for example-level hallucination and coverage error detection on four test sets (\S\ref{sec:test-sets}).
    }
    \label{tab:results all}
\end{table*}
\setlength{\belowcaptionskip}{0.0cm}

\subsection{Test Sets with Different Error Types}
\label{sec:test-sets}

Taking the annotations and synthetic errors generated above, we split the 72 ProCon topics into a train set (9 topics), development set (28 topics), and test set (35 topics).
We intentionally make our development and test sets substantially larger than our train set because our work focuses on evaluation (rather than training) of the NPOV Response Generator.
Our dataset contains two types of query-response pairs (paraphrased and unparaphrased) and three types of errors (synthetic full, organic full, and organic ambiguous). We evaluate the performance of our error detection methods on different slices of the test set to better understand where different approaches have strengths or weaknesses. Hence, each table in the results section states the specific test set slices evaluated:
\begin{itemize}[leftmargin=0.5cm]
\setlength\itemsep{0.0cm}
\item \textbf{Full organic}: unparaphrased error-free examples vs. organic full errors.
\item \textbf{Unparaphrased synthetic}: unparaphrased error-free examples vs. corresponding examples with synthetically-generated errors.
\item \textbf{Paraphrased synthetic}: paraphrased error-free examples vs. corresponding examples with synthetically-generated errors.
\item \textbf{Ambiguous organic}: unparaphrased error-free examples vs. ambiguous organic errors, including partial errors, repetition, and perspective confusion (\S\ref{sec:methods}).
\end{itemize}

\section{Results}
\label{sec:results}

\subsection{Example-Level Error Detection}
\label{ss:main_results}

First, we evaluate the three error detection methods (ROUGE, salience, and classifiers) at the example-level, i.e. detecting whether a query-response pair contains an error. The classifiers shown here are trained only on query-response pairs which are either error-free or contain \textit{synthetic} errors, including both paraphrased and unparaphrased versions (503 examples total); we explore the impact of training data on classifier performance in \S\ref{ss:classifier_ablation}.

Table~\ref{tab:results all} shows ROC AUC scores on the different test sets (\S\ref{sec:test-sets}) for all three methods.\footnote{The area under the receiver operating characteristic curve (ROC AUC) quantifies classification performance across classification thresholds by comparing the trade-off between true positive rate and false positive rate.} While the \textit{full organic} set (organic error-free examples vs. organic full errors) is the most realistic,  our synthetic sets allow for more controlled evaluations.
For all four test sets and for both hallucination and coverage errors, the ROC AUC difference when comparing the best performing method to either other method is statistically significant ($p < 0.001$), using the Wilcoxon statistic \citep{hanley-mcneil-1983} and Bonferroni correction for multiple comparisons \citep{bonferroni-1936-teoria,vanderweele-mathur-2019}.

Classifiers consistently outperform the other two methods by a large margin on all sets except ambiguous coverage errors (discussed below), with ROC AUCs above 90\% for both hallucination and coverage error detection, for all full error types (organic and synthetic, paraphrased and unparaphrased).
Comparing ROUGE and salience, results are mixed. On the full organic errors, ROUGE performs better at detecting hallucinations (84.0\% AUC), whereas salience performs better at detecting coverage errors (85.2\% AUC). 

For copy-like tasks with few expected word changes, ROUGE outperforms salience on both hallucination and coverage error detection (results on the unparaphrased synthetic errors set).
However, on the paraphrased synthetic errors, salience appears to capture the underlying semantics better than ROUGE, allowing it to more accurately detect both hallucination and coverage errors.

Finally, we evaluate our methods on ambiguous errors (including partial argument hallucination and coverage errors, argument repetition, and perspective confusion; see \S\ref{sec:methods}). ROUGE performs  well here, likely due to minimal natural paraphrasing from the NPOV Response Generator. Classifier ROC AUC scores drop substantially on ambiguous errors, likely because classifiers are trained only on full errors.  This discrepancy seems most problematic for coverage error detection, where classifiers perform even worse than ROUGE.
Future work should establish clearer definitions of ambiguous errors, allowing larger sets of ambiguous errors to be annotated and used to train classifiers.

\begin{table*}[t]
    \centering
    \begin{tabular}{|l|c|c|c|c||c|c|c|c|}
        \cline{2-9}
        \multicolumn{1}{c|}{} & \multicolumn{4}{c||}{Hallucinations} & \multicolumn{4}{c|}{Coverage Errors} \\
        \hline
        & Error-free   &    &   &  & Error-free   &    &   &  \\
        Test set error type & +Synth   & +Para & +Dev &+Org  & + Synth   & +Para & +Dev & +Org  \\
        \hline
        \hline
        Full organic             & 0.789 & 0.828 & 0.953 & 0.920 &  0.880 & 0.903 & 0.905 &  0.956 \\
        Ambiguous organic        & 0.807 & 0.820 & 0.851 & 0.862 & 0.702 & 0.529  & 0.756 & 0.640 \\
        \hline
    \end{tabular}
    \caption{ROC AUC scores for classifiers trained on different amounts and types of data (\S\ref{ss:classifier_ablation}), ordered from smallest to largest training set size.
    Table~\ref{tab:results all} results use the classifiers trained on +Dev.
    }
    \label{tab:classifier_ablation}
\end{table*}

\setlength{\belowcaptionskip}{-0.2cm}
\begin{table*}[t]
    \centering
    \begin{tabular}{|l|c|c||c|c|}
        \cline{2-5}
        \multicolumn{1}{c|}{} & \multicolumn{2}{c||}{Hallucinations} & \multicolumn{2}{c|}{Coverage Errors} \\
        \hline
        Test set error type & ROUGE & Salience & ROUGE & Salience \\
        \hline
        \hline
        Full organic & 0.673 & \textbf{0.724} & 0.669 & \textbf{0.799} \\
        Unparaphrased synthetic & 0.697 & \textbf{0.710} & 0.693 & \textbf{0.808} \\
        Paraphrased synthetic & 0.614 & \textbf{0.673} & 0.582 & \textbf{0.742} \\
        Ambiguous organic & \textbf{0.542} & \textbf{0.542} & 0.738 & \textbf{0.740} \\
        \hline
    \end{tabular}
    \caption{ROC AUC scores for word-level error detection using ROUGE and salience.}
    \label{tab:word-results}
\end{table*}
\setlength{\belowcaptionskip}{0.0cm}

\subsection{Classifier Training Data Ablations}
\label{ss:classifier_ablation}

We analyze the impact of different types and amounts of training data on classifier performance, considering the following four scenarios:
\begin{itemize}[leftmargin=0.5cm]
    \setlength\itemsep{0.0em}
    \item \textbf{Error-free +Synth}: all error-free query-response pairs, plus synthetic errors; training split only (70 examples).
    \item \textbf{+Para}: previous, plus equivalent paraphrased examples; training split only (138 examples).
    \item \textbf{+Dev}: previous, plus equivalent examples from the development split (503 examples).
    \item \textbf{+Org}: previous, plus examples with organic full errors; training and development splits (573 examples).
\end{itemize}
Table~\ref{tab:classifier_ablation} shows classifier performance on the full organic and ambiguous organic test sets (\S\ref{sec:test-sets}).
For coverage error detection, performance strictly improves on the full organic set as we add more training data. However, adding the organic error examples leads to a decline in performance on the ambiguous organic set.
For hallucination detection as well, we see performance improvement when adding more training data. Adding the organic errors leads to a performance drop on the full organic set, but not the ambiguous organic set.

Overall, adding more data, even consisting of synthetic errors, leads to improvements on most test sets for both hallucination and coverage error detection.
Surprisingly, adding organic errors on top leads to mixed results, showing that organic data is not necessarily always helpful or needed for good classifier performance. The \textbf{+Dev} scenario might already be large enough that the addition of organic errors does not provide benefit.

\subsection{Word-Level Error Detection}
\label{sec:word-results}

In practice, it may also be useful to locate specific response words that are hallucinated, or specific input words that are uncovered.
Of the methods in \S\ref{sec:methods}, ROUGE and salience both produce hallucination and coverage error scores at the word level. Specifically, the ROUGE coverage error score would be 0 if an input word is matched in the response (and 1 otherwise), and the ROUGE hallucination score would be 0 if a response word is matched in the input arguments (and 1 otherwise).
For salience, before example-level aggregation, scores are already computed per word (\S\ref{sec:salience}).
Sample word-level salience scores for hallucination and coverage errors are shown in Figure~\ref{fig:salience-map}.

We compare the word-level hallucination and coverage error scores from salience and ROUGE with the ground truth annotations of hallucinated and uncovered words annotated in our test sets (\S\ref{sec:annotation-procedure}).
Results are computed over all non-stop words in each test set, defining words by merging LLM tokens (\S\ref{sec:salience}); we compute results over all response words for hallucination word detection, and over all input words for coverage error word detection.
Results are reported in Table~\ref{tab:word-results}.
Differences between salience and ROUGE are statistically significant for all test sets and error types ($p < 0.001$) except the ambiguous organic error set, using the Wilcoxon statistic corrected for multiple comparisons as in \S\ref{ss:main_results}.
Salience performs equally to or better than ROUGE for detecting both hallucinated words in model responses and uncovered words in input arguments on all test sets.
On the test set with  paraphrased synthetic errors, salience has the largest relative gains over ROUGE, likely due to its ability to capture semantics even in cases of word mismatch, similar to the trends for example-level error detection.

\section{Discussion}

Overall, LLM-based classifiers trained on relatively small amounts of data perform surprisingly well, outperforming all other methods detecting full errors and obtaining promising ROC AUC scores between 90\% and 99\%. This is especially notable given that the classifiers are trained only on \textit{synthetic} hallucination and coverage errors and yet perform well on the organic test set.

While worse than the classifiers, the data-free methods presented here still achieve strong results.
Our experiments show that ROUGE is a strong data-free baseline for hallucination and coverage error detection in tasks with minimal paraphrasing. When more paraphrasing is expected, salience provides stronger results, appearing to better capture semantics than simple word matching.
Moreover, salience is effective for word-level hallucination and coverage error detection, allowing us to locate the parts of a generated response that are problematic.

Our experiments also show the value of different test set slices.
While the synthetically constructed datasets might diverge from the true data distribution, they offer a way to analyze strengths and weaknesses of different methods in an isolated fashion, e.g. paraphrased examples demonstrating the shortcomings of ROUGE. 

Finally, all methods struggle on ambiguous organic errors, although these results are inconclusive.
Largely, this set is a ``catch-all'' for problematic and low-agreement errors, possibly explaining the poor  performance of different error detection approaches. Training classifiers on this subset is important future work, but requires a larger dataset of more clearly-defined ambiguous errors.

\section{Related Work}

\paragraph{Errors in controlled text generation.}
Our approach to NPOV Response Generation using provided input perspectives is an example of \textit{retrieval augmented generation}, where information (e.g. a document or paragraph) is retrieved from a knowledge source (e.g. a search engine) and used to condition a model response \citep{li2022survey}. Like in our scenario, retrieval-augmented models sometimes exhibit hallucinations \cite{dziri2022} and coverage errors \cite{krishna2021} relative to the retrieved source.
Our error detection methods may be applied to these scenarios more generally.

Specifically, our task is closely related to \textit{table-to-text generation}, which aims to generate fluent and faithful natural language descriptions of tabular data. Table-to-text generation has been studied using a variety of datasets, including WikiBio \citep{wikibio2016}, ToTTo \citep{totto2020}, DART \citep{dart2021}, and WebNLG \citep{webnlg2017}.
Traditional metrics such as ROUGE, BLEU, and METEOR compare model responses to a reference output, but metrics developed specifically for table-to-text tasks (e.g. PARENT; \citealp{parent2019}) often consider both the table source and reference output when scoring a model response, to better preserve faithfulness to the source \citep{liu2021,thomson-reiter-2021-generation}.
Our work similarly compares model responses to the input source; however, our input fields are perspectives composed of several full sentences (arguments) rather than short expressions (e.g. entities or numbers that allow minimal paraphrasing, as in most table-to-text tasks).
For this reason, pure matching-based scoring approaches (e.g. ROUGE, BLEU, and PARENT) are less effective for our task.

More broadly, hallucinations are a common artifact in natural language generation (NLG). At a high level, they can be described as cases where generated output is ``unfaithful'' to provided or desired source content \cite{survey_hallucination}. Due to the fluency of modern NLG systems, hallucinations can remain undetected and mislead users. Tolerance to such errors is particularly low in summarization and table-to-text tasks, where a retrieved source is provided.
In the NPOV Response Task, we focus on \textit{full} errors, where a hallucinated or uncovered argument can be identified relatively unambiguously.

\vspace{-0.1cm}
\paragraph{Prompt-tuning.}
Both the NPOV Response Generator (\S\ref{ss:npov_writer}) and the  classifiers (\S\ref{ss:llm_classifiers}) use soft prompt-tuning, a method where only a small number of parameters are tuned and the base LLM is left unchanged~\cite{lester2021}.
\citet{mozes2023} show that LLMs can be prompt-tuned even on very small datasets to function as classifiers.
Open-source code to train such classifiers is available through the Gemma Responsible Generative AI Toolkit \citep{gemma-prompt-tuning}.

\vspace{-0.1cm}
\paragraph{Salience.}
Previous work has identified hallucinations in machine translation using proportions of source contributions to output tokens \cite{dale-etal-2023-detecting,voita-etal-2021-analyzing}, using aggregated layerwise token attribution \cite{ferrando2022}. 
Our salience-based method for error detection is similar, but  attributions are based on loss gradients~\cite{bastings-filippova-2020-elephant}.
We focus on dot products between gradients and inputs, which are often used to roughly quantify model attributions from input tokens \citep{ding-koehn-2021-evaluating,boggust2022beyond,zhao-etal-2022-really}.\footnote{We obtain similar results using gradient L2 norms.}
Previous work has applied gradient-based salience methods to fine-tuned encoder-decoder and encoder-only classification models \citep{tenney-etal-2020-language}. We extend this to decoder-only models, prompt-tuned on sequence-to-sequence tasks.

\section{Limitations}

Our work has several limitations.
The NPOV Response Generator is trained and evaluated only in English, and our NPOV Response Task does not address how to create the content in the perspectives and their arguments.
The arguments used in our work are pulled from ProCon, which limits both our set of controversial topics and our sets of perspectives (i.e. only pro and con; see Ethical Considerations); future work might consider more nuanced methods of perspective identification, selection, and/or generation.

Our work also does not focus on biases in LLM hallucinated or omitted content.
For example, the NPOV Response Generator may be more likely to hallucinate or omit arguments for specific topics or perspectives, e.g. based on the frequency of topics and perspectives in the LLM pre-training corpus \citep{durmus2023measuring}.
Even when focusing just on error detection rather than error content, we focus primarily on errors that are easy to identify and have high levels of inter-annotator agreement. Based on our own annotations, inter-annotator agreement on ambiguous errors is much lower than for full errors.
The majority of ambiguous errors that we observed can be classified as partial errors, repetition, or argument confusion (\S\ref{sec:methods}), but an important branch of future work is to establish more thorough taxonomies and annotation schemes for hallucination and coverage error types.

Finally, in future work we hope to evaluate whether our findings on LLM-based classifier performance generalize to other (ideally publicly available) LLMs. Many significant results involving LLMs have generalized to other LLMs (e.g. in-context learning, chain-of-thought reasoning, and parameter-efficient tuning methods; \citealp{brown2020language,wei2023chainofthought,lester2021}), but our results should be verified for other LLMs.

\section{Conclusion}
In this paper, we introduce the NPOV Response Task as an approach to retrieval augmented generation for controversial topics. We focus on response generation, after pro and con arguments are provided to an LLM.
We propose and evaluate methods for detecting hallucination and coverage errors in LLM-generated responses, and we demonstrate a synthetic error generation strategy that can be used to train and evaluate our proposed methods.
We find that prompt-tuned LLM classifiers trained only on synthetic errors achieve high error detection performance on organic examples. Our other methods, while performing worse than our classifiers, still achieve strong results without the need for training data.

\section*{Ethical Considerations}

With the rise of LLM-based chatbots and broader societal concerns about echo chambers, filter bubbles, and polarization, the ability of LLMs to provide neutral, factual, and nuanced responses to controversial topics is an important avenue of work. However, having LLMs respond to queries about controversial topics is inherently challenging: who decides what is controversial, neutral, and factual, and how this is encoded in an LLM is a hard and nebulous problem. Moreover, as LLMs and chatbot technologies become increasingly easy to create, maliciously engineered and maliciously applied models are likely to become more prevalent. Retrieval augmented generation is a way to control LLM responses in a maximally transparent way.

In this paper, we assume the existence of a database with NPOV-expressed perspectives. However, such a database is not an easy artifact to create, and the contents will often be hotly contested. The dataset we use is derived from Britannica's ProCon website \cite{procon}. However, this still reduces arguments to pro and con perspectives, which can reinforce a binary vision of the world. Our work also does not address how to best arrive at and reflect consensus on specific arguments. For example, when should the model express ``many experts'' vs. ``a few experts'' as a qualification for an argument? Failure here can serve to elevate fringe arguments. Even deciding whether a topic is controversial is already culturally charged. For instance, the subject of gun control might be a non-issue for some European countries yet remain polarizing in the United States. Similarly, omitting topics or arguments that are relevant for minorities or non-Western countries risks reinforcing systemic erasure and promoting socio-cultural biases. To address and mitigate these biases in a perspectives database, processes are necessary to ensure that the group of experts providing perspectives is diverse and multicultural.

The more basic question of when to apply an LLM in practical scenarios needs careful consideration. In some domains (e.g. medical information), even very low error rates may not be acceptable, while other domains (e.g. creative writing) have very different risk profiles. Proper evaluations, policies, and guardrails should be put in place before LLMs are applied in practice to new domains.

Finally, the computational footprints of the NPOV Response Generator and the LLM-based error classifiers are large, with each model built upon a 60B+ parameter LLM.
Similarly, computing salience maps for error detection requires computing gradients from the NPOV Response Generator itself, thus inducing a large computational cost.
Of the error detection methods evaluated in our work, ROUGE is by far the most computationally efficient. Future work may consider more computationally efficient approaches, such as evaluating smaller models as error detection classifiers.

\section*{Acknowledgements}
We would like to thank Ian Tenney, Jasmijn Bastings, Vinodkumar Prabhakaran, and the anonymous reviewers for valuable feedback.


\nocite{*}
\section{Bibliographical References}\label{sec:reference}
\bibliographystyle{lrec-coling2024-natbib}
\bibliography{custom}

\begin{thebibliography}{44}
\expandafter\ifx\csname natexlab\endcsname\relax\def\natexlab#1{#1}\fi

\bibitem[{Azure(2023)}]{azure-2023-retrieval}
Microsoft Azure. 2023.
\newblock \href
  {https://learn.microsoft.com/en-us/azure/search/retrieval-augmented-generation-overview}
  {Retrieval augmented generation ({RAG}) in {A}zure {C}ognitive {S}earch}.
\newblock \emph{Microsoft Azure Documentation}.

\bibitem[{Bastings and Filippova(2020)}]{bastings-filippova-2020-elephant}
Jasmijn Bastings and Katja Filippova. 2020.
\newblock \href {https://doi.org/10.18653/v1/2020.blackboxnlp-1.14} {The
  elephant in the interpretability room: Why use attention as explanation when
  we have saliency methods?}
\newblock In \emph{Proceedings of the Third BlackboxNLP Workshop on Analyzing
  and Interpreting Neural Networks for NLP}, pages 149--155, Online.
  Association for Computational Linguistics.

\bibitem[{Bird et~al.(2009)Bird, Klein, and Loper}]{nltk}
Steven Bird, Ewan Klein, and Edward Loper. 2009.
\newblock \href {https://www.nltk.org/} {\emph{Natural language processing with
  Python: {A}nalyzing text with the natural language toolkit}}.
\newblock O'Reilly Media.

\bibitem[{Boggust et~al.(2023)Boggust, Suresh, Strobelt, Guttag, and
  Satyanarayan}]{boggust2022beyond}
Angie Boggust, Harini Suresh, Hendrik Strobelt, John~V Guttag, and Arvind
  Satyanarayan. 2023.
\newblock \href {https://arxiv.org/abs/2206.02958} {Beyond faithfulness: A
  framework to characterize and compare saliency methods}.
\newblock In \emph{Proceedings of the ACM Conference on Fairness,
  Accountability, and Transparency (FAccT)}.

\bibitem[{Bonferroni(1936)}]{bonferroni-1936-teoria}
C.E. Bonferroni. 1936.
\newblock \emph{Teoria statistica delle classi e calcolo delle
  probabilit{\`a}}.
\newblock Pubblicazioni del R. Istituto superiore di scienze economiche e
  commerciali di Firenze. Seeber.

\bibitem[{Brown et~al.(2020)Brown, Mann, Ryder, Subbiah, Kaplan, Dhariwal,
  Neelakantan, Shyam, Sastry, Askell, Agarwal, Herbert-Voss, Krueger, Henighan,
  Child, Ramesh, Ziegler, Wu, Winter, Hesse, Chen, Sigler, Litwin, Gray, Chess,
  Clark, Berner, McCandlish, Radford, Sutskever, and
  Amodei}]{brown2020language}
Tom~B. Brown, Benjamin Mann, Nick Ryder, Melanie Subbiah, Jared Kaplan,
  Prafulla Dhariwal, Arvind Neelakantan, Pranav Shyam, Girish Sastry, Amanda
  Askell, Sandhini Agarwal, Ariel Herbert-Voss, Gretchen Krueger, Tom Henighan,
  Rewon Child, Aditya Ramesh, Daniel~M. Ziegler, Jeffrey Wu, Clemens Winter,
  Christopher Hesse, Mark Chen, Eric Sigler, Mateusz Litwin, Scott Gray,
  Benjamin Chess, Jack Clark, Christopher Berner, Sam McCandlish, Alec Radford,
  Ilya Sutskever, and Dario Amodei. 2020.
\newblock \href {https://arxiv.org/abs/2005.14165} {Language models are
  few-shot learners}.

\bibitem[{Chang and Bergen(2024)}]{chang-bergen-2023-language}
Tyler~A. Chang and Benjamin~K. Bergen. 2024.
\newblock \href {https://arxiv.org/abs/2303.11504} {Language model behavior: A
  comprehensive survey}.
\newblock \emph{Computational Linguistics}.

\bibitem[{Chowdhery et~al.(2023)Chowdhery, Narang, Devlin, Bosma, Mishra,
  Roberts, Barham, Chung, Sutton, Gehrmann, Schuh, Shi, Tsvyashchenko, Maynez,
  Rao, Barnes, Tay, Shazeer, Prabhakaran, Reif, Du, Hutchinson, Pope, Bradbury,
  Austin, Isard, Gur-Ari, Yin, Duke, Levskaya, Ghemawat, Dev, Michalewski,
  Garcia, Misra, Robinson, Fedus, Zhou, Ippolito, Luan, Lim, Zoph, Spiridonov,
  Sepassi, Dohan, Agrawal, Omernick, Dai, Pillai, Pellat, Lewkowycz, Moreira,
  Child, Polozov, Lee, Zhou, Wang, Saeta, Diaz, Firat, Catasta, Wei,
  Meier-Hellstern, Eck, Dean, Petrov, and Fiedel}]{palm-paper}
Aakanksha Chowdhery, Sharan Narang, Jacob Devlin, Maarten Bosma, Gaurav Mishra,
  Adam Roberts, Paul Barham, Hyung~Won Chung, Charles Sutton, Sebastian
  Gehrmann, Parker Schuh, Kensen Shi, Sasha Tsvyashchenko, Joshua Maynez,
  Abhishek Rao, Parker Barnes, Yi~Tay, Noam Shazeer, Vinodkumar Prabhakaran,
  Emily Reif, Nan Du, Ben Hutchinson, Reiner Pope, James Bradbury, Jacob
  Austin, Michael Isard, Guy Gur-Ari, Pengcheng Yin, Toju Duke, Anselm
  Levskaya, Sanjay Ghemawat, Sunipa Dev, Henryk Michalewski, Xavier Garcia,
  Vedant Misra, Kevin Robinson, Liam Fedus, Denny Zhou, Daphne Ippolito, David
  Luan, Hyeontaek Lim, Barret Zoph, Alexander Spiridonov, Ryan Sepassi, David
  Dohan, Shivani Agrawal, Mark Omernick, Andrew~M. Dai,
  Thanumalayan~Sankaranarayana Pillai, Marie Pellat, Aitor Lewkowycz, Erica
  Moreira, Rewon Child, Oleksandr Polozov, Katherine Lee, Zongwei Zhou, Xuezhi
  Wang, Brennan Saeta, Mark Diaz, Orhan Firat, Michele Catasta, Jason Wei,
  Kathy Meier-Hellstern, Douglas Eck, Jeff Dean, Slav Petrov, and Noah Fiedel.
  2023.
\newblock \href {https://arxiv.org/abs/2204.02311} {{PaLM}: {S}caling language
  modeling with {P}athways}.
\newblock \emph{Journal of Machine Learning Research}, 24(240):1--113.

\bibitem[{Chung et~al.(2022)Chung, Hou, Longpre, Zoph, Tay, Fedus, Li, Wang,
  Dehghani, Brahma et~al.}]{chung2022scaling}
Hyung~Won Chung, Le~Hou, Shayne Longpre, Barret Zoph, Yi~Tay, William Fedus,
  Eric Li, Xuezhi Wang, Mostafa Dehghani, Siddhartha Brahma, et~al. 2022.
\newblock \href {https://arxiv.org/abs/2210.11416} {Scaling
  instruction-finetuned language models}.
\newblock \emph{arXiv preprint}.

\bibitem[{Cohen et~al.(2022)Cohen, Roberts, Molina, Butryna, Jin, Kulshreshtha,
  Hutchinson, Zevenbergen, Aguera-Arcas, ching Chang, Cui, Du, Adiwardana,
  Chen, Lepikhin, Chi, Hoffman-John, Cheng, Lee, Krivokon, Qin, Hall, Fenton,
  Soraker, Meier-Hellstern, Olson, Aroyo, Bosma, Pickett, Menegali, Croak,
  Díaz, Lamm, Krikun, Morris, Shazeer, Le, Bernstein, Rajakumar, Kurzweil,
  Thoppilan, Zheng, Bos, Duke, Doshi, Zhao, Prabhakaran, Rusch, Li, Huang,
  Zhou, Xu, and Chen}]{lamda-paper}
Aaron~Daniel Cohen, Adam Roberts, Alejandra Molina, Alena Butryna, Alicia Jin,
  Apoorv Kulshreshtha, Ben Hutchinson, Ben Zevenbergen, Blaise~Hilary
  Aguera-Arcas, Chung ching Chang, Claire Cui, Cosmo Du, Daniel De~Freitas
  Adiwardana, Dehao Chen, Dmitry~(Dima) Lepikhin, Ed~H. Chi, Erin Hoffman-John,
  Heng-Tze Cheng, Hongrae Lee, Igor Krivokon, James Qin, Jamie Hall, Joe
  Fenton, Johnny Soraker, Kathy Meier-Hellstern, Kristen Olson, Lora~Mois
  Aroyo, Maarten~Paul Bosma, Marc~Joseph Pickett, Marcelo~Amorim Menegali,
  Marian Croak, Mark Díaz, Matthew Lamm, Maxim Krikun, Meredith~Ringel Morris,
  Noam Shazeer, Quoc~V. Le, Rachel Bernstein, Ravi Rajakumar, Ray Kurzweil,
  Romal Thoppilan, Steven Zheng, Taylor Bos, Toju Duke, Tulsee Doshi,
  Vincent~Y. Zhao, Vinodkumar Prabhakaran, Will Rusch, YaGuang Li, Yanping
  Huang, Yanqi Zhou, Yuanzhong Xu, and Zhifeng Chen. 2022.
\newblock \href {https://arxiv.org/abs/2201.08239} {{LaMDA}: {L}anguage models
  for dialog applications}.
\newblock \emph{arXiv preprint}.

\bibitem[{Dale et~al.(2023)Dale, Voita, Barrault, and
  Costa-juss{\`a}}]{dale-etal-2023-detecting}
David Dale, Elena Voita, Loic Barrault, and Marta~R. Costa-juss{\`a}. 2023.
\newblock \href {https://doi.org/10.18653/v1/2023.acl-long.3} {Detecting and
  mitigating hallucinations in machine translation: {M}odel internal workings
  alone do well, sentence similarity even better}.
\newblock In \emph{Proceedings of the 61st Annual Meeting of the Association
  for Computational Linguistics (Volume 1: Long Papers)}, pages 36--50.

\bibitem[{Denil et~al.(2014)Denil, Demiraj, and
  De~Freitas}]{denil-2014-extraction}
Misha Denil, Alban Demiraj, and Nando De~Freitas. 2014.
\newblock \href {https://arxiv.org/abs/1412.6815} {Extraction of salient
  sentences from labelled documents}.
\newblock \emph{arXiv preprint}.

\bibitem[{Dhingra et~al.(2019)Dhingra, Faruqui, Parikh, Chang, Das, and
  Cohen}]{parent2019}
Bhuwan Dhingra, Manaal Faruqui, Ankur Parikh, Ming-Wei Chang, Dipanjan Das, and
  William Cohen. 2019.
\newblock \href {https://aclanthology.org/P19-1483.pdf} {Handling divergent
  reference texts when evaluating table-to-text generation}.
\newblock In \emph{Proceedings of the 57th Annual Meeting of the Association
  for Computational Linguistics}, pages 4884--4895. Association for
  Computational Linguistics.

\bibitem[{Ding and Koehn(2021)}]{ding-koehn-2021-evaluating}
Shuoyang Ding and Philipp Koehn. 2021.
\newblock \href {https://doi.org/10.18653/v1/2021.naacl-main.399} {Evaluating
  saliency methods for neural language models}.
\newblock In \emph{Proceedings of the 2021 Conference of the North American
  Chapter of the Association for Computational Linguistics: Human Language
  Technologies}, pages 5034--5052, Online. Association for Computational
  Linguistics.

\bibitem[{Durmus et~al.(2023)Durmus, Nyugen, Liao, Schiefer, Askell, Bakhtin,
  Chen, Hatfield-Dodds, Hernandez, Joseph, Lovitt, McCandlish, Sikder, Tamkin,
  Thamkul, Kaplan, Clark, and Ganguli}]{durmus2023measuring}
Esin Durmus, Karina Nyugen, Thomas~I. Liao, Nicholas Schiefer, Amanda Askell,
  Anton Bakhtin, Carol Chen, Zac Hatfield-Dodds, Danny Hernandez, Nicholas
  Joseph, Liane Lovitt, Sam McCandlish, Orowa Sikder, Alex Tamkin, Janel
  Thamkul, Jared Kaplan, Jack Clark, and Deep Ganguli. 2023.
\newblock \href {https://arxiv.org/abs/2306.16388} {Towards measuring the
  representation of subjective global opinions in language models}.
\newblock \emph{arXiv preprint}.

\bibitem[{Dziri et~al.(2022)Dziri, Milton, Yu, Zaiane, and Reddy}]{dziri2022}
Nouha Dziri, Sivan Milton, Mo~Yu, Osmar Zaiane, and Siva Reddy. 2022.
\newblock \href {https://doi.org/10.18653/v1/2022.naacl-main.387} {On the
  origin of hallucinations in conversational models: Is it the datasets or the
  models?}
\newblock In \emph{Proceedings of the 2022 Conference of the North American
  Chapter of the Association for Computational Linguistics: Human Language
  Technologies}, pages 5271--5285.

\bibitem[{Ferrando et~al.(2022)Ferrando, G{\'a}llego, Alastruey, Escolano, and
  Costa-juss{\`a}}]{ferrando2022}
Javier Ferrando, Gerard~I. G{\'a}llego, Belen Alastruey, Carlos Escolano, and
  Marta~R. Costa-juss{\`a}. 2022.
\newblock \href {https://aclanthology.org/2022.emnlp-main.599/} {Towards
  opening the black box of neural machine translation: Source and target
  interpretations of the transformer}.
\newblock In \emph{Proceedings of the 2022 Conference on Empirical Methods in
  Natural Language Processing}, pages 8756--8769.

\bibitem[{Gardent et~al.(2017)Gardent, Shimorina, Narayan, and
  Perez-Beltrachini}]{webnlg2017}
Claire Gardent, Anastasia Shimorina, Shashi Narayan, and Laura
  Perez-Beltrachini. 2017.
\newblock \href {https://aclanthology.org/W17-3518/} {The {W}eb{NLG} challenge:
  Generating text from {RDF} data}.
\newblock In \emph{Proceedings of the 10th International Conference on Natural
  Language Generation}.

\bibitem[{Google(2024)}]{gemma-prompt-tuning}
Google. 2024.
\newblock \href {https://ai.google.dev/responsible/input_output} {Gemma:
  Responsible generative {AI} toolkit}.
\newblock \emph{Google AI for Developers}.

\bibitem[{Hanley and McNeil(1983)}]{hanley-mcneil-1983}
James~A Hanley and Barbara~J McNeil. 1983.
\newblock \href {https://pubmed.ncbi.nlm.nih.gov/6878708/} {A method of
  comparing the areas under receiver operating characteristic curves derived
  from the same cases.}
\newblock \emph{Radiology}, 148(3):839--843.

\bibitem[{Iyer and Thallam(2023)}]{iyer-thallam-2023-building}
Anand Iyer and Rajesh Thallam. 2023.
\newblock \href
  {https://cloud.google.com/blog/products/ai-machine-learning/generative-ai-applications-with-vertex-ai-palm-2-models-and-langchain}
  {Building generative {AI} applications made easy with {V}ertex {AI} {PaLM}
  {API} and {LangChain}}.
\newblock \emph{Google Cloud Blog, AI and Machine Learning}.

\bibitem[{Ji et~al.(2023)Ji, Lee, Frieske, Yu, Su, Xu, Ishii, Bang, Madotto,
  and Fung}]{survey_hallucination}
Ziwei Ji, Nayeon Lee, Rita Frieske, Tiezheng Yu, Dan Su, Yan Xu, Etsuko Ishii,
  Ye~Jin Bang, Andrea Madotto, and Pascale Fung. 2023.
\newblock \href {https://arxiv.org/abs/2202.03629} {Survey of hallucination in
  natural language generation}.
\newblock \emph{ACM Comput. Surv.}, 55(12).

\bibitem[{Krishna et~al.(2021)Krishna, Roy, and Iyyer}]{krishna2021}
Kalpesh Krishna, Aurko Roy, and Mohit Iyyer. 2021.
\newblock \href {https://aclanthology.org/2021.naacl-main.393/} {Hurdles to
  progress in long-form question answering}.
\newblock In \emph{Proceedings of the 2021 Conference of the North American
  Chapter of the Association for Computational Linguistics: Human Language
  Technologies}, pages 4940--4957.

\bibitem[{Lebret et~al.(2016)Lebret, Grangier, and Auli}]{wikibio2016}
R{\'e}mi Lebret, David Grangier, and Michael Auli. 2016.
\newblock \href {https://aclanthology.org/D16-1128/} {Neural text generation
  from structured data with application to the biography domain}.
\newblock In \emph{Proceedings of the 2016 Conference on Empirical Methods in
  Natural Language Processing}, pages 1203--1213. Association for Computational
  Linguistics.

\bibitem[{Lester et~al.(2021)Lester, Al-Rfou, and Constant}]{lester2021}
Brian Lester, Rami Al-Rfou, and Noah Constant. 2021.
\newblock \href {https://aclanthology.org/2021.emnlp-main.243.pdf} {The power
  of scale for parameter-efficient prompt tuning}.
\newblock In \emph{Proceedings of the 2021 Conference on Empirical Methods in
  Natural Language Processing}, pages 3045--3059. Association for Computational
  Linguistics.

\bibitem[{Lewis et~al.(2020)Lewis, Perez, Piktus, Petroni, Karpukhin, Goyal,
  K\"{u}ttler, Lewis, Yih, Rockt\"{a}schel, Riedel, and
  Kiela}]{lewis-retrieval-2020}
Patrick Lewis, Ethan Perez, Aleksandra Piktus, Fabio Petroni, Vladimir
  Karpukhin, Naman Goyal, Heinrich K\"{u}ttler, Mike Lewis, Wen-tau Yih, Tim
  Rockt\"{a}schel, Sebastian Riedel, and Douwe Kiela. 2020.
\newblock \href {https://arxiv.org/abs/2005.11401} {Retrieval-augmented
  generation for knowledge-intensive {NLP} tasks}.
\newblock In \emph{Proceedings of the 34th International Conference on Neural
  Information Processing Systems}.

\bibitem[{Li et~al.(2022)Li, Su, Cai, Wang, and Liu}]{li2022survey}
Huayang Li, Yixuan Su, Deng Cai, Yan Wang, and Lemao Liu. 2022.
\newblock \href {https://arxiv.org/abs/2202.01110} {A survey on
  retrieval-augmented text generation}.
\newblock \emph{arXiv preprint}.

\bibitem[{Li et~al.(2016)Li, Chen, Hovy, and
  Jurafsky}]{li-etal-2016-visualizing}
Jiwei Li, Xinlei Chen, Eduard Hovy, and Dan Jurafsky. 2016.
\newblock \href {https://doi.org/10.18653/v1/N16-1082} {Visualizing and
  understanding neural models in {NLP}}.
\newblock In \emph{Proceedings of the 2016 Conference of the North {A}merican
  Chapter of the Association for Computational Linguistics: Human Language
  Technologies}, pages 681--691, San Diego, California. Association for
  Computational Linguistics.

\bibitem[{Lin(2004)}]{lin-2004-rouge}
Chin-Yew Lin. 2004.
\newblock \href {https://aclanthology.org/W04-1013/} {{ROUGE}: A package for
  automatic evaluation of summaries}.
\newblock In \emph{Text Summarization Branches Out}, pages 74--81. Association
  for Computational Linguistics.

\bibitem[{Liu et~al.(2021)Liu, Zheng, Chang, and Sui}]{liu2021}
Tianyu Liu, Xin Zheng, Baobao Chang, and Zhifang Sui. 2021.
\newblock \href {https://arxiv.org/abs/2102.08585} {Towards faithfulness in
  open domain table-to-text generation from an entity-centric view}.
\newblock \emph{Proceedings of the AAAI Conference on Artificial Intelligence},
  35(15):13415--13423.

\bibitem[{Mozes et~al.(2023)Mozes, Hoffmann, Tomanek, Kouate, Thain, Yuan,
  Bolukbasi, and Dixon}]{mozes2023}
Maximilian Mozes, Jessica Hoffmann, Katrin Tomanek, Muhamed Kouate, Nithum
  Thain, Ann Yuan, Tolga Bolukbasi, and Lucas Dixon. 2023.
\newblock \href {https://doi.org/10.18653/v1/2023.findings-emnlp.30} {Towards
  agile text classifiers for everyone}.
\newblock In \emph{Findings of the Association for Computational Linguistics:
  EMNLP 2023}, pages 400--414, Singapore. Association for Computational
  Linguistics.

\bibitem[{Nan et~al.(2021)Nan, Radev, Zhang, Rau, Sivaprasad, Hsieh, Tang,
  Vyas, Verma, Krishna, Liu, Irwanto, Pan, Rahman, Zaidi, Mutuma, Tarabar,
  Gupta, Yu, Tan, Lin, Xiong, Socher, and Rajani}]{dart2021}
Linyong Nan, Dragomir Radev, Rui Zhang, Amrit Rau, Abhinand Sivaprasad,
  Chiachun Hsieh, Xiangru Tang, Aadit Vyas, Neha Verma, Pranav Krishna,
  Yangxiaokang Liu, Nadia Irwanto, Jessica Pan, Faiaz Rahman, Ahmad Zaidi,
  Mutethia Mutuma, Yasin Tarabar, Ankit Gupta, Tao Yu, Yi~Chern Tan,
  Xi~Victoria Lin, Caiming Xiong, Richard Socher, and Nazneen~Fatema Rajani.
  2021.
\newblock \href {https://doi.org/10.18653/v1/2021.naacl-main.37} {{DART}:
  Open-domain structured data record to text generation}.
\newblock In \emph{Proceedings of the 2021 Conference of the North American
  Chapter of the Association for Computational Linguistics: Human Language
  Technologies}, pages 432--447, Online. Association for Computational
  Linguistics.

\bibitem[{Parikh et~al.(2020)Parikh, Wang, Gehrmann, Faruqui, Dhingra, Yang,
  and Das}]{totto2020}
Ankur Parikh, Xuezhi Wang, Sebastian Gehrmann, Manaal Faruqui, Bhuwan Dhingra,
  Diyi Yang, and Dipanjan Das. 2020.
\newblock \href {https://doi.org/10.18653/v1/2020.emnlp-main.89} {{ToTTo}: A
  controlled table-to-text generation dataset}.
\newblock In \emph{Proceedings of the 2020 Conference on Empirical Methods in
  Natural Language Processing (EMNLP)}, pages 1173--1186, Online. Association
  for Computational Linguistics.

\bibitem[{{ProCon.org}(2022)}]{procon}
{ProCon.org}. 2022.
\newblock \url{https://www.procon.org/}.
\newblock Accessed: 2022-10-12.

\bibitem[{Sheng et~al.(2019)Sheng, Chang, Natarajan, and
  Peng}]{sheng-etal-2019-woman}
Emily Sheng, Kai-Wei Chang, Premkumar Natarajan, and Nanyun Peng. 2019.
\newblock \href {https://doi.org/10.18653/v1/D19-1339} {The woman worked as a
  babysitter: On biases in language generation}.
\newblock In \emph{Proceedings of the 2019 Conference on Empirical Methods in
  Natural Language Processing and the 9th International Joint Conference on
  Natural Language Processing (EMNLP-IJCNLP)}, pages 3407--3412, Hong Kong,
  China. Association for Computational Linguistics.

\bibitem[{Shuster et~al.(2021)Shuster, Poff, Chen, Kiela, and
  Weston}]{shuster-etal-2021-retrieval-augmentation}
Kurt Shuster, Spencer Poff, Moya Chen, Douwe Kiela, and Jason Weston. 2021.
\newblock \href {https://doi.org/10.18653/v1/2021.findings-emnlp.320}
  {Retrieval augmentation reduces hallucination in conversation}.
\newblock In \emph{Findings of the Association for Computational Linguistics:
  EMNLP 2021}, pages 3784--3803.

\bibitem[{Tenney et~al.(2020)Tenney, Wexler, Bastings, Bolukbasi, Coenen,
  Gehrmann, Jiang, Pushkarna, Radebaugh, Reif, and
  Yuan}]{tenney-etal-2020-language}
Ian Tenney, James Wexler, Jasmijn Bastings, Tolga Bolukbasi, Andy Coenen,
  Sebastian Gehrmann, Ellen Jiang, Mahima Pushkarna, Carey Radebaugh, Emily
  Reif, and Ann Yuan. 2020.
\newblock \href {https://doi.org/10.18653/v1/2020.emnlp-demos.15} {The language
  interpretability tool: Extensible, interactive visualizations and analysis
  for {NLP} models}.
\newblock In \emph{Proceedings of the 2020 Conference on Empirical Methods in
  Natural Language Processing: System Demonstrations}, pages 107--118, Online.
  Association for Computational Linguistics.

\bibitem[{Thomson and Reiter(2021)}]{thomson-reiter-2021-generation}
Craig Thomson and Ehud Reiter. 2021.
\newblock \href {https://aclanthology.org/2021.inlg-1.23} {Generation
  challenges: Results of the accuracy evaluation shared task}.
\newblock In \emph{Proceedings of the 14th International Conference on Natural
  Language Generation}, pages 240--248. Association for Computational
  Linguistics.

\bibitem[{VanderWeele and Mathur(2019)}]{vanderweele-mathur-2019}
Tyler~J. VanderWeele and Maya~B. Mathur. 2019.
\newblock \href {https://www.ncbi.nlm.nih.gov/pmc/articles/PMC6395159/} {Some
  desirable properties of the bonferroni correction: {I}s the bonferroni
  correction really so bad?}
\newblock \emph{American Journal of Epidemiology}, 188(3):617--618.

\bibitem[{Voita et~al.(2021)Voita, Sennrich, and
  Titov}]{voita-etal-2021-analyzing}
Elena Voita, Rico Sennrich, and Ivan Titov. 2021.
\newblock \href {https://doi.org/10.18653/v1/2021.acl-long.91} {Analyzing the
  source and target contributions to predictions in neural machine
  translation}.
\newblock In \emph{Proceedings of the 59th Annual Meeting of the Association
  for Computational Linguistics and the 11th International Joint Conference on
  Natural Language Processing (Volume 1: Long Papers)}, pages 1126--1140.

\bibitem[{Wei et~al.(2022)Wei, Wang, Schuurmans, Bosma, Ichter, Xia, Chi, Le,
  and Zhou}]{wei2023chainofthought}
Jason Wei, Xuezhi Wang, Dale Schuurmans, Maarten Bosma, Brian Ichter, Fei Xia,
  Ed~Chi, Quoc Le, and Denny Zhou. 2022.
\newblock \href {https://arxiv.org/abs/2201.11903} {Chain-of-thought prompting
  elicits reasoning in large language models}.
\newblock In \emph{Proceedings of the 36th International Conference on Neural
  Information Processing Systems}.

\bibitem[{Wulczyn et~al.(2017)Wulczyn, Thain, and Dixon}]{wulczyn2017exmachina}
Ellery Wulczyn, Nithum Thain, and Lucas Dixon. 2017.
\newblock \href {https://doi.org/10.1145/3038912.3052591} {Ex machina: Personal
  attacks seen at scale}.
\newblock In \emph{Proceedings of the 26th International Conference on World
  Wide Web}, WWW '17, page 1391–1399, Republic and Canton of Geneva, CHE.
  International World Wide Web Conferences Steering Committee.

\bibitem[{Zhang et~al.(2020)Zhang, Kishore, Wu, Weinberger, and
  Artzi}]{zhang2020bertscore}
Tianyi Zhang, Varsha Kishore, Felix Wu, Kilian~Q. Weinberger, and Yoav Artzi.
  2020.
\newblock \href {https://arxiv.org/abs/1904.09675} {{BERTScore}: Evaluating
  text generation with {BERT}}.
\newblock In \emph{International Conference on Learning Representations}.

\bibitem[{Zhao et~al.(2022)Zhao, Yuanzhe, Zhongtao, Yiming, Jun, and
  Kang}]{zhao-etal-2022-really}
Yang Zhao, Zhang Yuanzhe, Jiang Zhongtao, Ju~Yiming, Zhao Jun, and Liu Kang.
  2022.
\newblock \href {https://aclanthology.org/2022.ccl-1.82} {Can we really trust
  explanations? {E}valuating the stability of feature attribution explanation
  methods via adversarial attack}.
\newblock In \emph{Proceedings of the 21st Chinese National Conference on
  Computational Linguistics}, pages 932--944, Nanchang, China. Chinese
  Information Processing Society of China.

\end{thebibliography}

\appendix

\section*{Appendices}
\section{ProCon Dataset Details}

\begin{table*}[ht]
    \centering
    \begin{tabular}{|p{0.1\linewidth}|p{0.1\linewidth}|p{0.7\linewidth}|}
         \hline
         Split & \# of topics & Topics  \\
         \hline
         Train & 9 & \textit{Animal Dissection; Concealed Handguns; Cuba Embargo; Filibuster; Free College; GMOs (Genetically Modified Organisms); Net Neutrality; Obesity; Vaping E-Cigarettes} \\
         \hline
         Dev & 28 & \textit{Binge-Watching; Cancel Culture; Churches and Taxes; College Education; Corporal Punishment; Daylight Saving Time; Dress Codes; Electoral College; Employer Vaccine Mandates; Fighting in Hockey; Golf; Homework; Kneeling during National Anthem; Marijuana (CBD) for Pets; Olympics; Penny; Pit Bull Bans; Pokémon; School Vouchers; Space Colonization; Standardized Tests; Student Loan Debt; Tablets vs. Textbooks; Teacher Tenure; Uber \& Lyft; US Supreme Court Packing; Video Games and Violence; Zoos} \\         
         \hline
         Test & 35 & \textit{Abortion; American Socialism; Animal Testing; Artificial Intelligence; Banned Books; Bottled Water Ban; Cell Phone Radiation; Climate Change; Corporate Tax Rate; DACA \& Dreamers; DC and Puerto Rico Statehood; Defund the Police; Drone Strikes Overseas; Fracking; Gold Standard; Gun Control; Historic Statue Removal; Mandatory National Service; Minimum Wage; OTC Birth Control; Paying College Athletes; Police Body Cameras; Prescription Drug Costs; Private Prisons; Recreational Marijuana Legalization; Reparations for Slavery; Right to Health Care; Sanctuary Cities; Saturday Halloween; School Uniforms; Social Media; Social Security Privatization; Universal Basic Income; Vaccines for Kids; Vegetarianism} \\
        \hline
    \end{tabular}
    \caption{ProCon topics assigned to the different dataset splits.}
    \label{tab:procon_splits}
\end{table*}

We use the perspectives and arguments for the different topics listed on Britannica's ProCon website as of October 2022 \cite{procon}. We randomly split the 72 ProCon topics into train, dev, and test, as shown in Table~\ref{tab:procon_splits}, ensuring no overlap in topics across these splits. In line with ProCon's usage guidelines, all arguments are used verbatim as stated on the specific topic website under the section ``Pro \& Con Arguments''. We scrape the subtitles of the pro and con columns as our arguments. The median number of arguments per pro and con perspective per topic is 4, with a maximum of 23 and a minimum of 2.
The ProCon data is publicly available through their website, containing no personally-identifying information about individuals.
We follow the guidelines specified by ProCon on ``How to Use'' their data (\url{https://www.procon.org/faqs/#II}).

\section{Prompt-Tuning Details}
\label{app:prompt-tuning}

This section discusses implementation details of (1) the NPOV Response Generator and (2) the hallucination and coverage error classifiers, which are both based on prompt-tuning an LLM. We use the same prompt-tuning settings for both. 

We deliberately refrain from resource-intense hyperparameter tuning and instead use configurations previously shown to work well \cite{mozes2023}:
we use soft prompt lengths of 5 tokens initialized with a random sample of the model's 5K most frequent token vocabulary embeddings \cite{lester2021}; we then train with a learning rate of $0.1$ with 500 warm-up steps and linear decay, using small batch sizes of 16 for training and limiting training to 20K steps. In most cases, we reach the maximum development set performance after 2-5K steps.
Prompt-tuning runs take a maximum of 4 hours per run on 64 TPUv4 chips.

\begin{figure*}[ht]
    \centering
    \includegraphics[width=\linewidth]{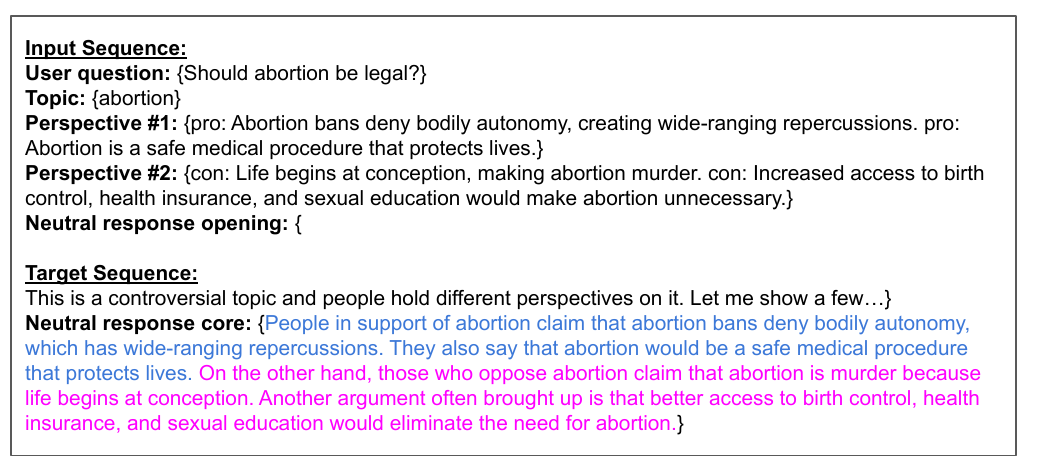} 
    \caption{Task format for the NPOV Response Task.}
    \label{fig:nuanced_response_representation}
\end{figure*}

\begin{figure*}[ht]
    \centering
    \includegraphics[width=\linewidth]{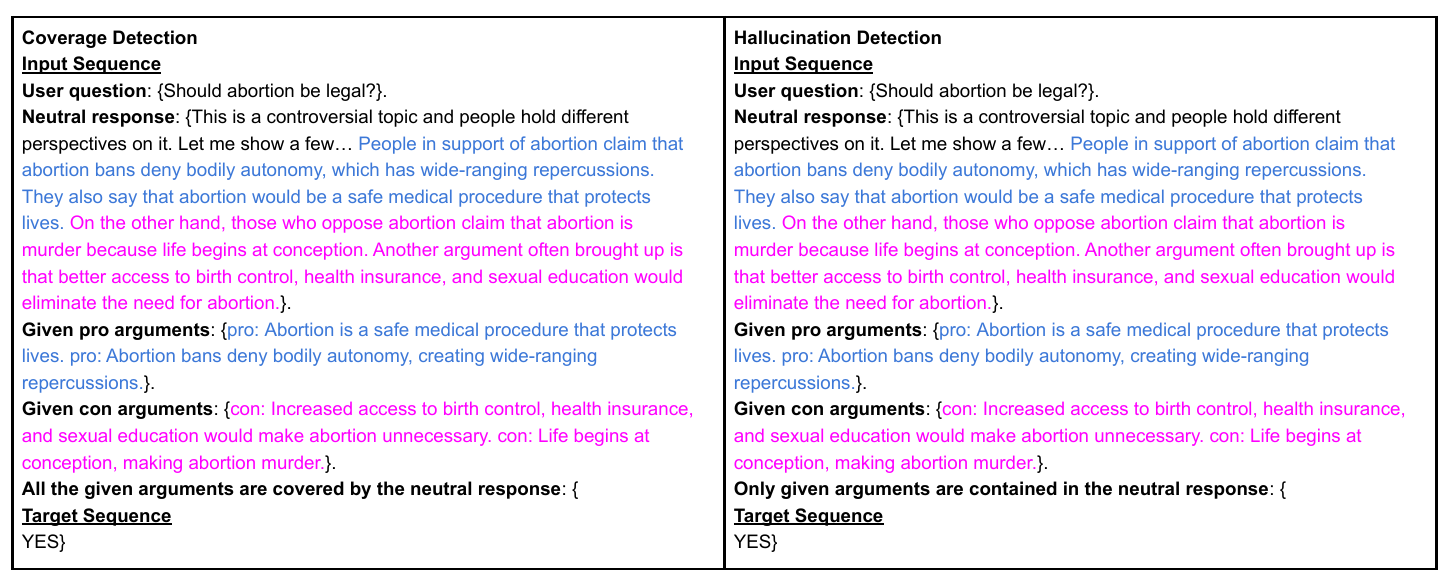}
    \caption{Task format for LLM-based error classifiers.}
    \label{fig:llm_representation}
\end{figure*}

For the task representations, we utilize a ``curly braces format'' to verbalize the task, consisting of several key-value pairs in the input and target sequence for the LLM. This format is easily picked up by modern LLMs, as they have typically been exposed to code during pre-training.
Figure~\ref{fig:nuanced_response_representation} shows how we format the task for the NPOV Response Task (\S\ref{ss:npov_writer}).
Figure~\ref{fig:llm_representation} shows how we format the error classification tasks (\S\ref{ss:llm_classifiers}).

\subsection{Classifier Ablation: Annotation-Free Scenario}
As an additional experiment, we analyze whether we can obtain good classifiers for hallucination and coverage error detection by just re-utilizing the original training data from the NPOV Response Task, without the need to perform any of the manual annotations described in \S\ref{sec:annotation-procedure}.
We turn the data used to train the NPOV Response Generator into error classifier training data by (1) treating NPOV Response Task training examples as no error-examples, and (2) adding synthetic errors according to our procedure in \S\ref{sec:synthetic-generation}. We call this approach ``annotation-free'' because we do not have to obtain any additional human annotations for classifier training.
The resulting hallucination and coverage error classifiers are trained on 50 error-free examples and 131 examples with synthetic errors.

Table~\ref{tab:anno_free_classifier} shows results on the organic test sets for the ``annotation-free'' classifiers. Overall, these results are significantly worse than results with the non-annotation-free classifiers (compare to Table~\ref{tab:classifier_ablation}), and often worse than other data-free approaches (compare to ROUGE and salience in Table~\ref{tab:results all}). This suggests that error classifier training may require organic model responses, even if the errors are synthetically generated.

\begin{table}[ht]
    \centering
    \begin{tabular}{|l|c|c|}
        \hline
        Test Set & Hallucination & Coverage \\
        \hline
        Full organic & 0.739 & 0.896 \\
        Ambiguous org.  & 0.732 & 0.804 \\
        \hline
    \end{tabular}
    \caption{Annotation-free classifer error detection ROC AUC scores.}
    \label{tab:anno_free_classifier}
\end{table}

\section{Salience Formulas}
\label{app:salience-formulas}
In \S\ref{sec:salience}, we describe how we compute a word-to-word salience map $M_{\textrm{words}} \in \R^{m \times n}$, where $m$ is the number of non-stop words in the input arguments and $n$ is the number of non-stop words in the generated NPOV response.
Our salience maps are based on gradient times input attribution scores, but we obtain comparable results using gradient L2 norms.
Here, we include formal equations defining our hallucination and coverage error detection metrics based on $M_{\textrm{words}}$.

Assume $I_{\textrm{pro}}$ and $I_{\textrm{con}}$ are the lists of non-stop words in the input pro and con arguments respectively.
Assume $O_{\textrm{resp}}$ is the list of non-stop words in the generated NPOV main response.
For each input word $w_i \in I_{\textrm{pro}} \cup I_{\textrm{con}}$, we define its contribution score $\alpha_i$ as its maximum contribution to any response word (i.e. the maximum across the corresponding row of $M_{\textrm{words}}$):
\begin{equation}
\label{eq:app-contrib}
\alpha_i = \textrm{max}(M_{\textrm{words}}[i, :])
\end{equation}
Similarly, for each output word $w_j \in O_{\textrm{resp}}$, we define its attribution score $\beta_j$ as its maximum attribution from any input argument word (i.e. the maximum across the corresponding column of $M_{\textrm{words}}$):
\begin{equation}
\label{eq:app-attrib}
\beta_j = \textrm{max}(M_{\textrm{words}}[:, j])
\end{equation}
Sample contribution and attribution scores for input words and response words respectively are shown in Figure~\ref{fig:salience-map}.
For word-level error detection (\S\ref{sec:word-results}), these word-level scores can be converted into coverage error scores $1.0 - \alpha_i$ and hallucination scores $1.0 - \beta_j$.

For example-level error detection (\S\ref{ss:main_results}), we compute an example-level coverage error score by (1) taking the geometric mean of word-level contribution scores for each input perspective, (2) taking the minimum of the two perspective scores (to reflect the fact that both perspectives must contribute), and (3) subtracting from 1.0 (lower contributions are more likely to be coverage errors):
\begin{align*}
    s_{\textrm{cov}} = 1.0 - \textrm{min}\Big(&\textrm{gmean}_{w_i \in I_{\textrm{pro}}}(\alpha_i), \\
    & \textrm{gmean}_{w_i \in I_{\textrm{con}}}(\alpha_i)\Big)
\end{align*}
We compute an example-level hallucination score by (1) taking the geometric mean of word-level attribution scores in the NPOV main response, and (2) subtracting from 1.0 (lower attributions are more likely to be hallucinations):
\[ s_{\textrm{hall}} = 1.0 - \textrm{gmean}_{w_j \in O_{\textrm{resp}}}(\beta_j) \]
Note that $s_{\textrm{cov}}, s_{\textrm{hall}} \in [0, 1]$ because entries of $M_{\textrm{words}}$ are in $[0, 1]$.
We evaluate these hallucination and coverage error scores for example-level error detection in \S\ref{ss:main_results}.

\subsection{Alternative Salience Aggregation Methods}
Above, we use the maximum function (in Equations \ref{eq:app-contrib} and \ref{eq:app-attrib}) to aggregate a contribution score for each input word and an attribution score for each response word.
This is based on the observation that covered input words tend to have a high contribution to at least one response word, and non-hallucinated response words tend to have a high attribution from at least one input argument word. In Table \ref{tab:alternate-salience}, we report ROC AUC results on the full organic test set using different methods to aggregate word contributions and attributions in Equations \ref{eq:app-contrib} and \ref{eq:app-attrib}.
Specifically, we consider (1) the sum (i.e. the sum over all response contributions for each input word to quantify coverage, and the sum over all input attributions for each response word to quantify non-hallucination), and (2) the (negative) entropy.
Lower entropies indicate less distributed contributions/attributions, such as when most of the contribution/attribution is to/from a single word (a pattern which appears in the majority of covered and non-hallucinated words).

We find that entropies perform worse than the maximum and sum aggregation functions for both hallucination and coverage error detection.
The sum performs best for hallucination detection (summing input attributions for each response word), but the maximum performs best for coverage error detection (taking the maximum response contribution for each input word).
We use the maximum in the main results for consistency and to avoid overfitting to the test set.

\begin{table}[t]
    \centering
    \begin{tabular}{|l|c|c|}
        \hline
        Aggregation & Hallucination & Coverage \\
        \hline
        Max & 0.808 & \textbf{0.852} \\
        Sum & \textbf{0.846} & 0.809 \\
        Negative entropy & 0.786 & 0.664 \\
        \hline
    \end{tabular}
    \caption{Example-level error detection ROC AUC scores for salience using different methods to aggregate a contribution score for each input word (coverage) and an attribution score for each response word (hallucination).}
    \label{tab:alternate-salience}
\end{table}

\section{Human Annotation Details}
\label{app:annotations}
For the human annotations in \S\ref{ss:interannotator-agreement}, our annotation service provider was paid 49 USD per hour for a total of 25 hours of work; they state that they ensure fair payment to annotators.
The 10 annotators were specialized workers in the United States contracted by our annotation provider.
Our annotation provider reported self-disclosed genders and age brackets of annotators, but this information was not used in our analyses.
Our annotations focused on attributes of our NPOV Response Generator query-response pairs, collecting annotation labels but no other data generated by the annotators.
To reduce annotation bias, annotators were not told how the labeled examples would be used, and they were not told that the response was machine-generated.

\end{document}